\documentclass[conference]{IEEEtran}

\usepackage[utf8]{inputenc}
\usepackage{cite}
\usepackage[linesnumbered,vlined,ruled,boxed,commentsnumbered]{algorithm2e}
\usepackage{algpseudocode}
\usepackage{bm} 
\usepackage{subfig}

\usepackage{amsmath,amssymb,amsfonts}
\usepackage{commath}
\usepackage{float}

\usepackage[flushleft]{threeparttable}
\usepackage{cleveref}

\usepackage{graphics,graphicx}
\usepackage{subfloat}

\usepackage[acronym,smallcaps]{glossaries}

\usepackage{url}            

\newcommand{\fref}[1]{Fig. \ref{#1}}
\newcommand{\tref}[1]{Table \ref{#1}}

\newcommand{\sref}[1]{Section \ref{#1}}

\usepackage{xcolor} %

\newacronym{larfdssom}{LARFDSSOM}{\textit{Local
Adaptive Receptive Field Dimension Selective Self-organizing Map}}
\newacronym{soms}{SOMs}{\textit{Self-Organizing Maps}}
\newacronym{som}{SOM}{\textit{Self-Organizing Map}}
\newacronym{larfsom}{LARFSOM}{\textit{Local Adaptive Receptive Field Self-Organizing Map}}
\newacronym{dssom}{DSSOM}{\textit{Dimension Selective Self-Organizing Map}}
\newacronym{sssom}{SS-SOM}{\textit{Semi-Supervised Self-Organizing Map}}
\newacronym{bsssom}{Batch SS-SOM}{\textit{Batch Semi-Supervised Self-Organizing Map}}
\newacronym{ssl}{SSL}{\textit{Semi-Supervised Learning}}
\newacronym{dssl}{DSSL}{\textit{Deep Semi-Supervised Learning}}
\newacronym{lvq}{LVQ}{\textit{Learning Vector Quantization}}
\newacronym{s3vm}{$S^3VM$}{\textit{Semi-supervised Support Vector Machines}}
\newacronym{svm}{SVM}{\textit{Support Vector Machines}}
\newacronym{hmrf}{HMRFs}{\textit{Hidden Markov Random Fields}}
\newacronym{cnn}{CNN}{\textit{Convolutional Neural Networks}}
\newacronym{em}{EM}{\textit{Expectation Maximization}}
\newacronym{lp}{LP}{\textit{Label Propagation}}
\newacronym{ijcnn}{IJCNN}{\textit{International Joint Conference on Neural Networks}}
\newacronym{mlp}{MLP}{\textit{Multilayer Perceptron}}
\newacronym{grlvq}{GRLVQ}{\textit{Generalized Relevance Learning Vector Quantization}}
\newacronym{ls}{LS}{\textit{Label Spreading}}
\newacronym{lhs}{LHS}{\textit{Latin Hypercube Sampling}}
\newacronym{csom}{CSOM}{\textit{Convolutional Self-Organizing Map}}
\newacronym{dsom}{DSOM}{\textit{Deep Self-Organizing Map}}
\newacronym{gpu}{GPU}{\textit{Graphics Processing Units}}
\newacronym{svhn}{SVHN}{\textit{Street View House Numbers}}

\begin{document}

\title{Deep Categorization with Semi-Supervised Self-Organizing Maps}
\author{
  \IEEEauthorblockN{Pedro H. M. Braga, \textit{Member}, \textit{IEEE}, Heitor R. Medeiros, \textit{Member}, \textit{IEEE},\\Hansenclever F. Bassani, \textit{Member}, \textit{IEEE}}
  \IEEEauthorblockA{Center of Informatics - CIn, Universidade Federal de Pernambuco, Recife, PE, Brazil, 50.740-560\\
  Email: \{phmb4, hrm, hfb\}@cin.ufpe.br}
}

\maketitle

\begin{abstract}
Nowadays, with the advance of technology, there is an increasing amount of unstructured data being generated every day. However, it is a painful job to label and organize it. Labeling is an expensive, time-consuming, and difficult task. It is usually done manually, which collaborates with the incorporation of noise and errors to the data. Hence, it is of great importance to developing intelligent models that can benefit from both labeled and unlabeled data. Currently, works on unsupervised and semi-supervised learning are still being overshadowed by the successes of purely supervised learning. However, it is expected that they become far more important in the longer term. This article presents a semi-supervised model, called Batch Semi-Supervised Self-Organizing Map (Batch SS-SOM), which is an extension of a SOM incorporating some advances that came with the rise of Deep Learning, such as batch training. The results show that Batch SS-SOM is a good option for semi-supervised classification and clustering. It performs well in terms of accuracy and clustering error,  even with a small number of labeled samples, as well as when presented to unsupervised data, and shows competitive results in transfer learning scenarios in traditional image classification benchmark datasets. 
\end{abstract}

\IEEEpeerreviewmaketitle

\section{Introduction}


Nowadays, with the advance of technology, there is a plentiful amount of unstructured data available. However, organize and label them is considerably challenging. Labeling is an expensive, time-consuming, and difficult task that is usually done manually. People can label with different formats and styles, incorporating noise and errors to the dataset \cite{jindal2016learning}. For instance, competitions like Kaggle: ImageNet Object Localization Challenge \cite{kaggle}, tries to encourage this kind of practice in order to obtain more reliable and bigger datasets continuously. There, participants are challenged in tasks to identifying objects within an image, so those images can then be further classified and annotated to be incorporated into datasets.

It is well-known that supervised learning algorithms normally reach good performances when high amounts of reliable and properly labeled data are available \cite{zhang2018survey}. On the other hand, \gls{ssl} purpose is to categorize (classify or cluster) data even with a lack of properly labeled examples. To this extent, \gls{ssl} algorithms put forward learning approaches that benefit from both labeled or unlabeled data.

Still, the abundant unlabeled data has a large amount of discriminating information that can be fully explored by \gls{ssl} algorithms and then combined with the prior information available from the smaller number of labeled samples. In this context, previous works in \gls{ssl} have contributed directly to a variety of areas in different application scenarios, such as traffic classification \cite{erman2007offline}, health monitoring \cite{longstaff2010improving}, and person identification \cite{balcan2005person}.

Moreover, those datasets are most common in the form of a high number of dimensions, then providing complex data structures to be fed to the models. Such high-dimensional data space imposes great challenges for the traditional machine learning approaches because they normally have the presence of noisy or uncorrelated data in some dimensions. Furthermore, due to the curse of dimensionality, traditional distance metrics may become meaningless, making objects appear to be approximately equidistant from each other. So, many approaches have been applied to deal with this problem. For example, \gls{lvq} and \gls{som} based models, such as  \cite{hammer2002generalized, bassani2015larfdssom}.

Therefore, not only the datasets itself can be used for clustering or classification tasks, but also its characteristics (features), or learned representations, that can be found and extracted using deep learning models. Then, it can be transferred and fed to independent learning models using transfer learning techniques \cite{oliver2018realistic}, which is a common practice employed to workaround the computational cost necessary to train huge datasets while also exploring generalization capabilities. For instance, it is very common to see many works using pre-trained features of ImageNet \cite{imagenet}. However, such strategies are often neglected in \gls{ssl}, but it is still a good starting point, which can be further explored.

In this article, we propose a new model called \gls{bsssom}, which is a novel approach to the previous \gls{sssom} model that can be easily scaled to a wide range of deep learning tasks. To achieve this, many modifications were incorporated into the baseline model to allow dealing with batches of samples and easily couple the proposed model in Deep Learning architectures. For the evaluation, we compared it with other semi-supervised models as well as studied its performance and behavior under different amounts of available labels in a variety of deep learning benchmark datasets.

The rest of this article is organizing as follows: \sref{sec:background} presents a short background related to the areas in which this paper is inserted. \sref{sec:sssom} discusses the \gls{sssom}, the baseline model for the current work. \sref{sec:batch-sssom} describes in detail the proposed method. \sref{sec:experiments} presents the experiments, methodology, the obtained results, and comparisons. Finally, \sref{sec:conclusions} debates the obtained results and concludes this paper, as well as indicates future directions.

\section{Background}
\label{sec:background}



\begin{figure}[ht]
	\centering
	\includegraphics[width=1\linewidth]{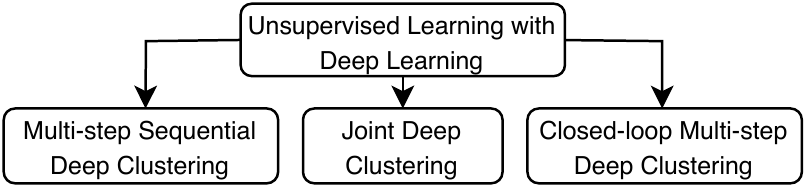}
	\caption{\textmd{(Deep) Unsupervised  Learning  Taxonomy \cite{nutakki2019introduction}.}}
	\label{fig:deep-clustering-taxonomy}
\end{figure}

According to \cite{deep-learning}, it is expected that unsupervised and semi-supervised learning becomes far more important in the longer term. On considering a purely unsupervised scenario, many approaches based on deep learning have been proposed recently. In that sense, \cite{nutakki2019introduction} divides them into three different main strategies, as illustrated in \fref{fig:deep-clustering-taxonomy}. 

The so-called Multi-Step Sequential Deep Clustering consists of two main steps: 1) learn a richer deep representation (also known as latent representation) of the input data; 2) perform clustering on this deep or latent representation. For instance, it can be distinguished by the use of transfer learning techniques \cite{oliver2018realistic}, relying on the use of pre-trained models to create or extract the representations that can be further fed to clustering models. The current paper is based on this approach. In Joint Deep Clustering, the step where the representation is learned is tightly coupled with the clustering. Hence, models are trained with a combined or joint loss function that favors learning a good representation while performing the clustering task itself. The Closed-loop Multi-step Deep Clustering is similar to Multi-Step Sequential Deep Clustering. However, after pre-training, the steps alternate in an iterative loop, where the output of the clustering method can be used to allow retraining or fine-tuning of the deep representation.


\begin{figure*}[H]
	\centering
	\includegraphics[width=0.8\linewidth]{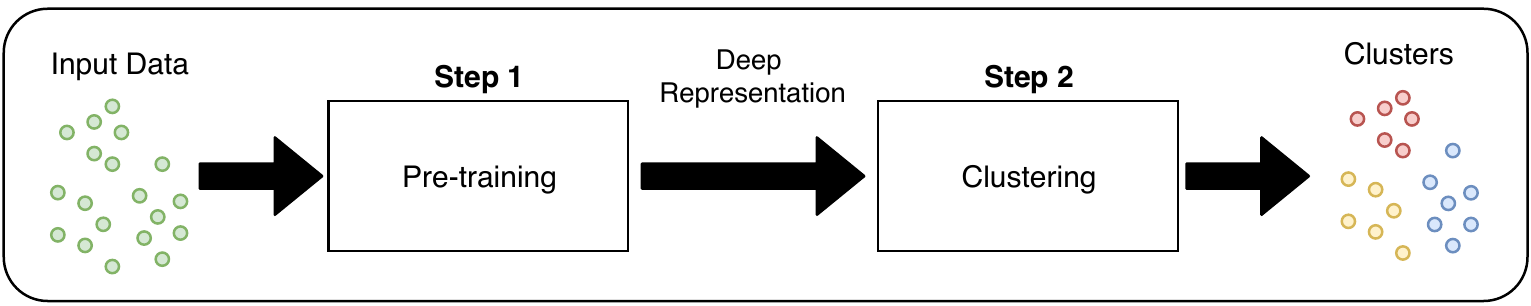}
	\caption{\textmd{Multi-Step Sequential Deep Clustering. In the first step the input date is used to train a model to create a deep representation. After that, this representation can be used in the second step by a clustering method to perform cluster.}}
	\label{fig:multi-step-sequential-deep-clustering}
\end{figure*}

On the other hand, Projected Clustering, Soft Projected Clustering, Subspace Clustering, and Hybrid algorithms are common approaches for the semi-supervised and unsupervised traditional context. They use diverse kinds of models, ranging from Prototype-based models algorithms and \gls{hmrf} to \gls{lp} \cite{schwenker2014pattern}.

\gls{ssl} can be further classified into semi-supervised classification and semi-supervised clustering \cite{schwenker2014pattern}. In semi-supervised classification, the training set is normally given in two parts: $S = \{(\textbf{x}_i, y_i) | \textbf{x}_i \in \mathbb{R}^d, y_i \in Y, 1 \leq i \leq M \}$ and $U = \{ \textbf{u}_i \in  \mathbb{R}^d | i = 1, \cdots, M \}$. Where \textit{S} and \textit{U} are the labeled and unlabeled data, respectively. It is possible to consider a traditional supervised scenario using only \textit{S} to build a classifier. However, the unsupervised estimation of the probability function \textit{p}(\textbf{x}) can take advantage of both \textit{S} and \textit{U}. Besides, classification tasks can reach a higher performance through the use of \gls{ssl} as a combination of supervised and unsupervised learning \cite{schwenker2014pattern}. 

Nonetheless, in the semi-supervised clustering, the aim is to group the data in an unknown number of groups relying on some kind of similarity or distance measures in combination with objective functions. Moreover, the nature of the data can make the clustering tasks challenging, so any kind of additional prior information can be useful to obtain a better performance. Therefore, the general idea behind semi-supervised clustering is to integrate some type of prior information in the process.

Many models from both kinds of approaches have been proposed over the years \cite{zhu2006semi}. However, as mentioned, conventional forms of clustering suffer when dealing with high-dimensional spaces. In this sense, \gls{som}-based algorithms have been proposed \cite{bassani2015larfdssom, braga2018semi, csom, braga2019semi}. However, most of them do not have any form to explore the benefits of more advanced techniques, even a simple form of mini-batch learning. The \gls{sssom} is explained in more detail in the next section in order to establish the ideas of the model proposed in this paper.

Finally, notice that \gls{ssl} is growing in machine learning alongside the deep learning context, as it is possible to see in \cite{kingma2014semi,chen2018semi,zhu2003semi,medeiros2019dynamic}. So, it is not unusual to find the term \gls{dssl} to express deep learning methods applicable to \gls{ssl}, as well as approaches combining more traditional models to work in a deep learning scenario. They range from approaches based on generative models \cite{kingma2014semi} to transfer learning \cite{oliver2018realistic}, convolutional, and SOM-based approaches \cite{csom,dsom}.

\section{SS-SOM}
\label{sec:sssom}

\gls{sssom} \cite{braga2018semi} is a semi-supervised \gls{som}, based on \gls{larfdssom} \cite{bassani2015larfdssom}, with a time-varying structure \cite{araujo2013self} and two different ways of learning. It is composed of a set of neurons (nodes) connected to form a map in which each node is a prototype representing a subset of input data. The nodes in \gls{sssom} can consider different relevances for the input dimensions and adapt its receptive field during the self-organization process. To do so, \gls{sssom} computes the called relevance vectors by estimating the average distance of each node to the input pattern that it clusters. The distance vectors are updated through a moving average of the observed distance between the input patterns and the current center vector (prototype). 

The \gls{sssom} can switch between a supervised or unsupervised learning procedure during the self-organization process according to the availability of the information about the class label for each input pattern. It modifies the \gls{larfdssom} to include concepts from the standard \gls{lvq} \cite{kohonen1995learning} when the class label of an input pattern is given. The general operation of the map consists of three phases. They are: 1) Organization; 2) Convergence; and 3) Clustering and/or Classification.


In the organization phase, after initialization, the nodes compete to form clusters of randomly chosen input patterns. There are two different ways to decide which node is the winner of a competition, which nodes need to be updated, and when a new node needs to be inserted. If the class label of the input pattern is provided, the supervised learning mode is used, and each winner node in the map will be associated with a respective class label. Otherwise, the unsupervised mode is employed.

When executing the unsupervised mode, given an unlabeled input pattern, the \gls{sssom} algorithm looks for a winner node disregarding their class labels. Therefore, the winner of a competition is the node that is the most activated according to a radial basis function with the receptive field adjusted as a function of its relevance vector. Also, the neighborhood of \gls{sssom} is formed by connecting nodes with others of the same class label, or with unlabeled nodes. Moreover, in \gls{sssom}, any node that does not win at least a minimum percentage of competitions will be removed from the map.


The convergence is pretty similar to the organization phase. However, there is no insertion of new nodes. After finishing the convergence phase, the map can cluster and classify input patterns. Depending on the amount and distribution of labeled input patterns presented to the network during the training, after the convergence phase, the map may have: 1) all the nodes labeled; 2) some nodes labeled; 3) no nodes labeled. All of these before mentioned situations are handled differently by \gls{sssom}. In \cite{braga2018semi}, a full description concerning each of them is given.


\section{Batch SS-SOM}
\label{sec:batch-sssom}

\begin{figure*}[ht]
	\centering
	\includegraphics[width=\linewidth]{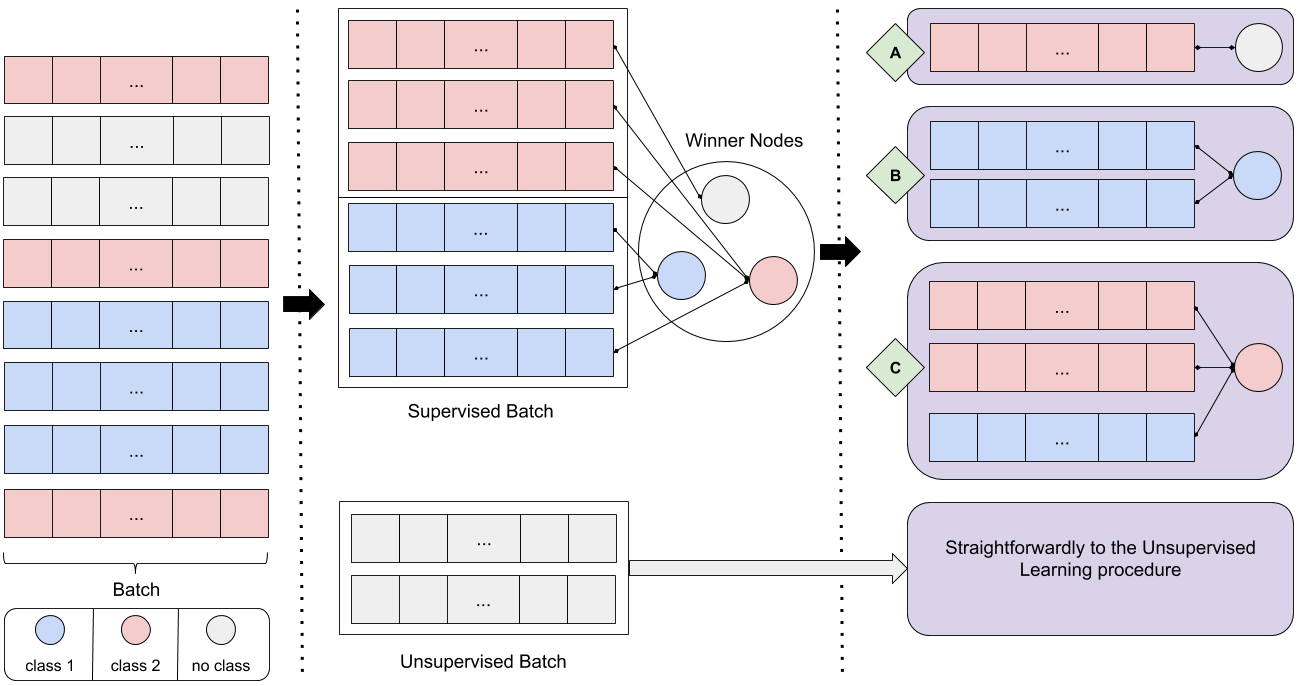}
	\caption{\textmd{The basic operation performed by \gls{bsssom} when a mini-batch is given, and its resulting cases.}}
	\label{fig:batch-sssom-1}
\end{figure*}

To extend the range of applications of \gls{sssom}, Batch \gls{sssom}\footnote{Available at: \url{https://github.com/phbraga/batch-sssom}} is introduced. Initially, to take advantage of \gls{gpu}, to allow mini-batch training, and thus to be more integrated with other Deep Learning approaches that commonly use the same framework and structure, the implementation uses the PyTorch framework. Moreover, three important modifications to the baseline model in order to improve its performance under the new set of conditions are proposed. 

First, when a mini-batch is given to the model, it is separated into two different mini-batches: 1) the unsupervised mini-batch; and 2) the supervised mini-batch, as shown by the first two columns of \fref{fig:batch-sssom-1}. For the unsupervised case, the key-point modification is to compute an average vector $\boldsymbol{\overline{X}}_u$ of all unlabeled samples that each winner node $j$ succeeded to be the most activated during the competition. After that, the process continues straightforwardly to the unsupervised procedure by sending all the average vectors and their representative winner node. 

On the other hand, the supervised scenario results in three distinct situations, as illustrated by the last column of \fref{fig:batch-sssom-1}, that must be handled differently after finding the winner node for each sample contained in the supervised mini-batch (likewise in \gls{sssom}):

\begin{enumerate}
\item \textbf{\fref{fig:batch-sssom-1}-A:} A node with an undefined class is the winner for a labeled sample;
\item \textbf{\fref{fig:batch-sssom-1}-B:} A node with a defined class is the winner for one or more samples of the same class;
\item \textbf{\fref{fig:batch-sssom-1}-C:} A node with a defined class is the winner for one or more samples of different classes, including or not its own.
\end{enumerate}

\begin{figure*}[ht]
	\centering
	\includegraphics[width=\linewidth]{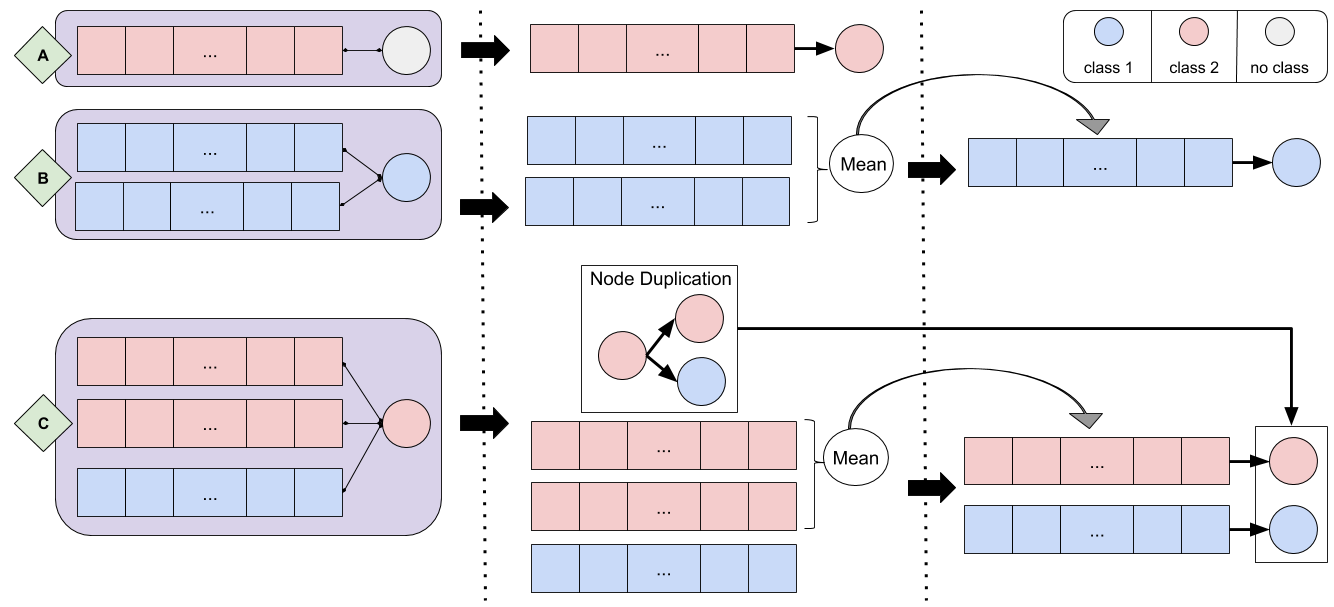}
	\caption{\textmd{How to handle each distinct situation from the \gls{bsssom} operation when a mini-batch is given.}}
	\label{fig:batch-sssom-2}
\end{figure*}

\fref{fig:batch-sssom-2} shows how each of these mentioned situations is handled. In \fref{fig:batch-sssom-2}-A workflow, the actions are to set the node class to be the same as the input pattern and then update its position towards such an input. 

Second, in \fref{fig:batch-sssom-2}-B, it is necessary to compute the average vector $\boldsymbol{\overline{X}}_l$, where $l$ is the related class label,  considering all the samples that are under this situation. Notice that $l$ is unique. Following, the usual supervised update procedure of \gls{sssom} is called, where the class is the same for both node and average sample vector.


Third, the case illustrated in \fref{fig:batch-sssom-2}-C is handled as follows: for all the classes contained in this subset of samples, every different class duplicates the original winner node $j$ by preserving the centroid vector $\boldsymbol{c}_j$, the distance vector $\boldsymbol{\delta}_j$, and the relevance vector $\boldsymbol{\omega}_j$, but setting the class of the new duplicated node to be the same as the current treated class, as well as setting its number of victories to zero. After that, for each class \textit{l} found in the current subset, a vector $\boldsymbol{\overline{X}}_l$ is calculated, and the respective duplicated node is updated using both $\boldsymbol{\overline{X}}_l$ and \textit{l}, suchlike in \fref{fig:batch-sssom-2}-B, in which the original winner node is updated using its corresponding vector and class.

Still, notice that when this situation occurs for an unlabeled winner node \textit{j}, the first calculated $\boldsymbol{\overline{X}}_u$ vector and its related class are used to update \textit{j} as in \fref{fig:batch-sssom-2}-A. Finally, all the operations executed in the \gls{bsssom} are performed in parallel on the \gls{gpu}, which optimizes the computational cost and allows the model to be applied to more complex tasks, datasets, and architectures.

\section{Experiments}
\label{sec:experiments}

The experiments were divided into two distinct scenarios. The first one is focused on comparing \gls{bsssom} with semi-supervised methods widely used in clustering tasks to show how competitive the model is. The latter demonstrates the capability of the proposed model to cluster and deal with extracted features from \gls{cnn} architectures.

\subsection{Parameters Sampling}

In order to properly adjust the parameters of the model, the \gls{lhs} \cite{helton2005comparison} was used. It is a statistical method for generating a random sample of parameter values from a multidimensional distribution. In this sense, for the first experimental scenario, we gathered 500 different parameter settings, i.e., the range of each parameter was divided into 500 intervals of equal probability to be sampled \cite{helton2005comparison}. For the latter scenario, we sampled 10 different parameter sets using \gls{lhs}. For both, a batch size of 32 was used. In \tref{tab:sssom-params}, the parameter ranges for the Batch \gls{sssom} is provided.

\begin{table}[ht!]
\centering
\begin{threeparttable}
\renewcommand{\arraystretch}{1.3}
\caption{Parameter Ranges of \gls{bsssom}}
\label{tab:sssom-params}
\centering
\begin{tabular}{lcc}
\hline
\bfseries  Parameters & \bfseries min & \bfseries max\\
\hline
Activation threshold ($a_t$) & 0.90 & 0.999 \\
Lowest cluster percentage (lp) & 0.001 & 0.01 \\
Relevance rate ($\beta$) & 0.001 & 0.5 \\
Max competitions ($age\_wins$) & $1 \times S^*$ & $100 \times S^*$ \\
Winner learning rate ($e_b$) & 0.001 & 0.2 \\
Wrong winner learning rate ($e_w$) & $0.01 \times e_b$ & $1 \times e_b$ \\
Neighbors learning rate ($e_n$) & $0.002 \times e_b$ & $ 1 \times e_b$ \\
Relevance smoothness ($s$) & 0.01 & 0.1 \\
Connection threshold ($minwd$) & 0 & 0.5 \\
Number of epochs ($epochs$) & 1 & 100 \\
\hline
\end{tabular}
\begin{tablenotes}
\small\item * \textit{S} is the number of input patterns in the dataset.
\end{tablenotes}
\end{threeparttable}
\end{table}

\subsection{Datasets}
Before underlying in more detail the two sets of experiments, it is important to specify the used datasets.


\subsubsection{UCI Datasets}

We select some datasets from the UCI machine learning repository \cite{uci-ml} that were previously used in similar works to perform a comparison with our approach in the same experimental setup used for them. They are Breast, Diabetes, Glass, Liver, Shape, and Vowel.

\subsubsection{MNIST}

\begin{figure}[H]
	\centering
	\includegraphics[width=\linewidth]{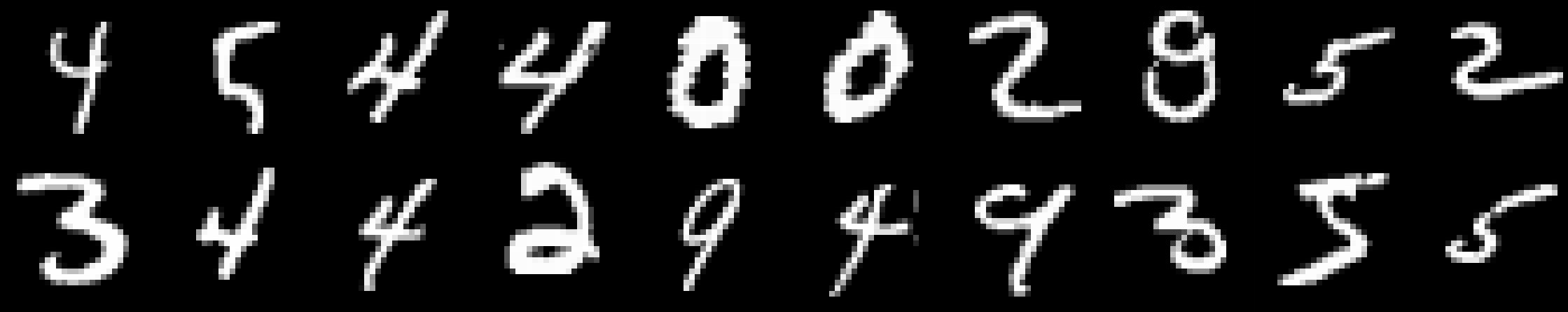}
	\caption{\textmd{MNIST samples.}}
	\label{fig:mnist-samples}
\end{figure}

The MNIST is a widely used image benchmark dataset of handwritten digits; it has 60,000 examples of the training set and 10,000 examples of the test set. Each sample fits into a 28x28 grayscale level bounding box~\cite{lecun1998mnist}. \fref{fig:mnist-samples} shows some of its samples. 

\subsubsection{Fashion-MNIST}

\begin{figure}[H]
	\centering
	\includegraphics[width=\linewidth]{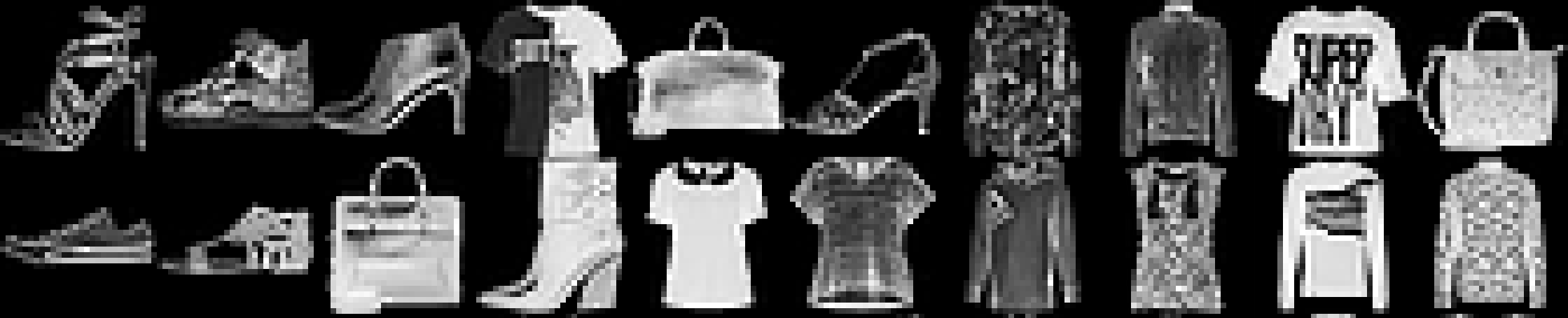}
	\caption{\textmd{Fashion-MNIST samples.}}
	\label{fig:fashion-mnist-samples}
\end{figure}

Fashion-MNIST is a fashion product dataset of Zalando's article images \cite{fashion}; it shares the same image size and structure of train and test splits of MNIST dataset. \fref{fig:fashion-mnist-samples} shows Fashion-MNIST samples.

\subsubsection{SVHN}

\begin{figure}[H]
	\centering
	\includegraphics[width=\linewidth]{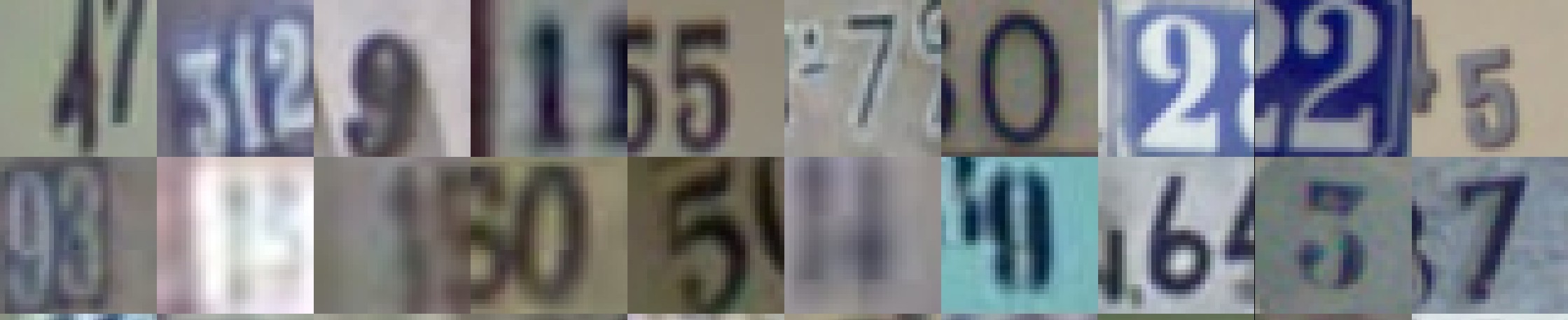}
	\caption{\textmd{SVHN samples.}}
	\label{fig:svhn-samples}
\end{figure}

The \gls{svhn} dataset is a real-world house numbers dataset obtained from Google Street View images. The \gls{svhn} is much harder than MNIST because images have a lack of contrast, normalization, and sometimes the digit has been overlapped by others, or it has noisy features. It consists of $73,257$ digits for training, $26,032$ digits for testing, and $531,131$ extra training data ranging from $0$ to $9$ digits\cite{svhn}. \fref{fig:svhn-samples} shows \gls{svhn} samples.

\subsection{\gls{bsssom} on UCI Datasets}

\tref{tab:results_uci} shows the best values of the Clustering Error (CE) of the models over 500 runs on each of them, and as high the value is, the better is the model. In addition to the comparisons of SOM-based models, DOC \cite{procopiuc2002monte} and PROCLUS \cite{aggarwal1999fast} models were used. They both are commonly used benchmarks for this type of datasets.

\gls{bsssom} showed to perform well in clustering tasks over a variety of datasets from UCI. In the Breast dataset, it achieved the same value as other clustering methods. In Diabetes and Vowel, it was statistically equal to LARFDSSOM~\cite{bassani2015larfdssom} / \gls{sssom}~\cite{braga2018semi} and ALT-SSSOM~\cite{braga2019semi}, respectively, because in the unlabeled scenario, it behaves similarly. In the Glass, Liver, and Shape datasets, the batch size has a slightly negative influence on the outcome, once it presented a small degradation in terms of performance, which is an effect of the mean vector update rule. However, the \gls{bsssom} accelerates the training process by using the mini-batches and can also be employed as a last layer of deep learning models to perform categorization tasks. In particular, it is important to point out that \gls{sssom} works exactly as LARFDSSOM when no labels are available.

\begin{table*}[ht]
\small
\centering
\renewcommand{\arraystretch}{1.3}
\caption{CE Results for Real-World Datasets. Best results for each dataset are shown in bold.}
\label{tab:results_uci}
\centering
\begin{tabular}{c||lllllll|cc}
\hline
\bfseries  CE & \bfseries Breast & \bfseries Diabetes & \bfseries Glass & \bfseries Liver & \bfseries Shape & \bfseries Vowel\\
\hline\hline
DOC \cite{procopiuc2002monte} & \textbf{0.763} & 0.654 & 0.439 & 0.580 & 0.419 & 0.142 \\
PROCLUS \cite{aggarwal1999fast}& 0.702 & 0.647 & 0.528 & 0.565 & 0.706 & 0.253 \\
LARFDSSOM \cite{bassani2015larfdssom} / SS-SOM \cite{braga2018semi} & \textbf{0.763}  & \textbf{0.727} & \textbf{0.575} & 0.580 & 0.719 & 0.317 \\
ALT-SSSOM \cite{braga2019semi} & \textbf{0.763}  & 0.697 & \textbf{0.575} & \textbf{0.603} & \textbf{0.738} & \textbf{0.319} \\
Batch SS-SOM & \textbf{0.763}  & 0.723 & 0.537 & 0.580 & 0.693 & 0.301  \\

\hline
\end{tabular}
\end{table*}

\subsection{\gls{bsssom} using Features Extracted from Custom \gls{cnn} Models}

Since \gls{bsssom} showed good results in comparison with its competitors in the first scenario, we accessed its performance in a more challenging task with high-dimensional data, such as images, using high-level features. To do so, we develop the following strategy. First, we trained a \gls{cnn} model from scratch and then extracted the features. More specifically, we extracted the features before the classifier layer, using them as input to \gls{bsssom}. Second, we defined several supervision rates, i.e., the percentage of available labels. It is worth mentioning that the sampling was not balanced. Also, this experiment indicates the effects of the number of labeled samples in the outcome results for MNIST, Fashion-MNIST, and SVHN. For this scenario, we started from MNIST and then expanded to the other datasets to guide a case study about the behavior of the model. The main idea is not to surpass any other model, but understand its behaviors when applied to more complex data structures or representations.

\begin{figure}[H]
    \centering
    \subfloat[Custom layer block for MNIST.]{\includegraphics[width=.95\linewidth]{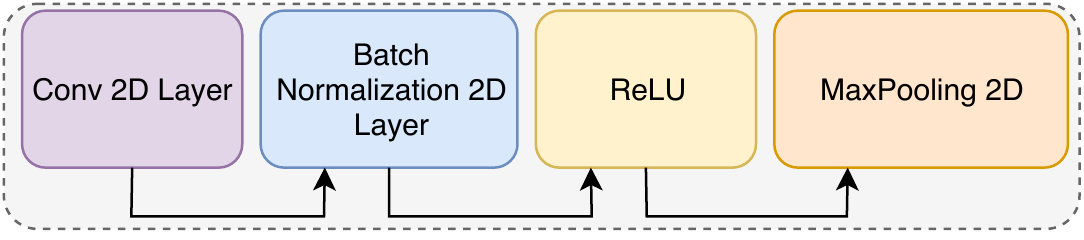}
    \label{fig:mnist_cnn_model_a}
    }
    
    \centering
    \subfloat[MNIST \gls{cnn} Model: Two custom layer blocks (\fref{fig:mnist_cnn_model_a}) followed by two fully-connected (dense) layers.]{\includegraphics[width=.95\linewidth]{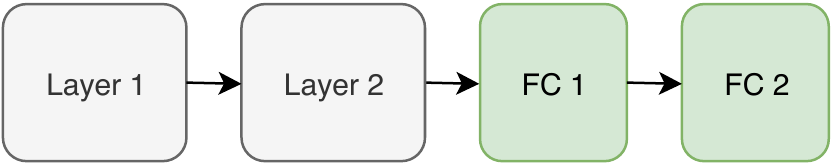} 
     \label{fig:mnist_cnn_model_b}
    }
    
    \centering
    \subfloat[\gls{bsssom} training pipeline: The previous FC2 is removed, the features are extracted from FC1 and then fed to \gls{bsssom}.]{\includegraphics[width=.95\linewidth]{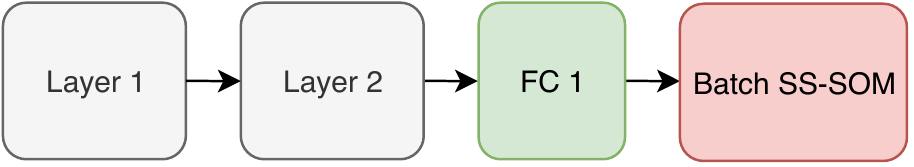}
    \label{fig:mnist_cnn_model_c}
    }
    \caption{MNIST Training Pipeline.}%
    \label{fig:mnist_cnn_model}%
\end{figure}

Different \gls{cnn} architectures were evaluated in order to achieve better results for each dataset in particular. For MNIST, \fref{fig:mnist_cnn_model_a} describes the \gls{cnn} layer block (convolution, batch normalization, ReLU and Max Pooling); \fref{fig:mnist_cnn_model_b} illustrates the full architecture: layer1 (16 filters, kernel size 5x5, stride 1x1, padding 2x2), layer2 (32 filters, kernel size 5x5, stride 1x1, padding 2x2), fully-connected 1 (FC1 with 32 neurons), fully-connected 2 (FC2 with 10 neurons); In \fref{fig:mnist_cnn_model_c}, we described the \gls{bsssom} training pipeline, in which we removed FC2 and extracted features to feed \gls{bsssom} with 32 input dimensions.

\begin{figure}[H]
    \centering
    \subfloat[Custom layer block for Fashion-MNIST.]{{\includegraphics[width=.95\linewidth]{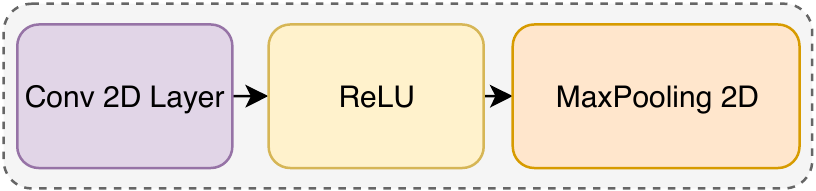}}
    \label{fig:fashion_mnist_cnn_model_a}
    }

    \subfloat[Fashion-MNIST CNN Model: Two custom layers (\fref{fig:fashion_mnist_cnn_model_a}), and a dropout layer followed by 3 fully-connected (dense) layers.]{{\includegraphics[width=.95\linewidth]{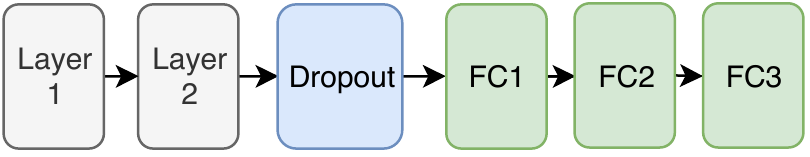} }
    \label{fig:fashion_mnist_cnn_model_b}
    }
    
    \subfloat[\gls{bsssom} training pipeline: The previous FC3 is removed, the features are extracted from FC2 and then fed to \gls{bsssom}.]{{\includegraphics[width=.95\linewidth]{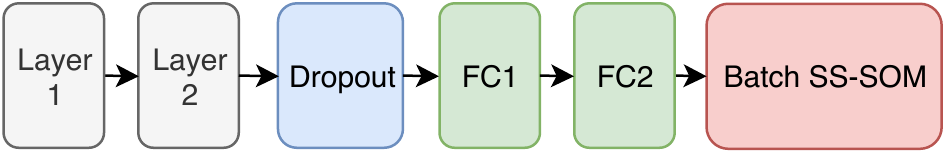} }
    \label{fig:fashion_mnist_cnn_model_c}
    }
    \caption{Fashion-MNIST Training Pipeline.}%
    \label{fig:fashion-mnist-cnn-model}%
\end{figure}

For Fashion-MNIST, \fref{fig:fashion_mnist_cnn_model_a} describes the \gls{cnn} layer block (convolution, ReLU and Max Pooling); In \fref{fig:fashion_mnist_cnn_model_b}, the full architecture is given: layer1 (64 filters, kernel size 5x5, stride 1x1), layer2 (32 filters, kernel size 5x5, stride 1x1), dropout (0.5), fully-connected 1 (FC1 with 128 neurons), fully-connected 2 (FC2 with 32 neurons), and fully-connected 3 (FC3 with 10 neurons); Still, \fref{fig:fashion_mnist_cnn_model_c} shows the \gls{bsssom} training pipeline, where we remove FC3 layer to extract the features to \gls{bsssom} with 32 dimensions.

\begin{figure}[H]
    \centering
    \subfloat[SVHN \gls{cnn} Model: One  convolutional 2D layer followed by a max-pooling 2D, other convolutional 2D layer and 3 fully-connected (dense) layers.]{{\includegraphics[width=.95\linewidth]{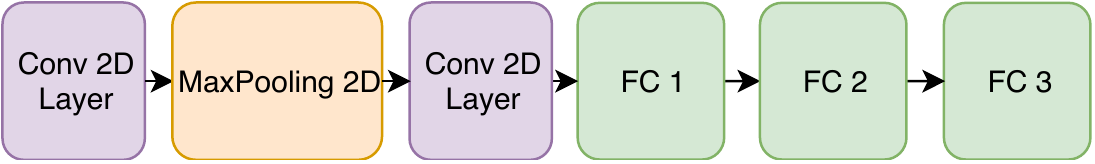}}
    \label{fig:svhn_cnn_model_a}
    }

    \subfloat[\gls{bsssom} training pipeline: The previous FC3 is removed, the features are extracted from FC2 and then fed to \gls{bsssom}.]{{\includegraphics[width=.95\linewidth]{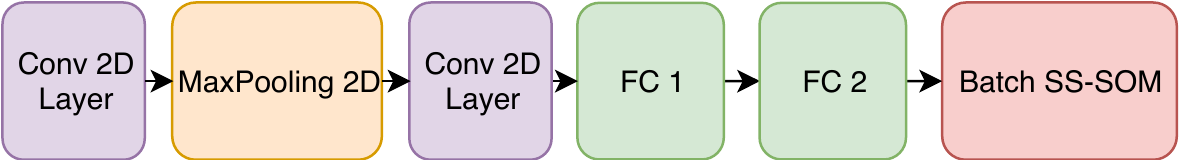}}
    \label{fig:svhn_cnn_model_b}
    }
    
    \caption{SVHN Training Pipeline.}%
    \label{fig:svhn_cnn_model}
\end{figure}

Lastly, for \gls{svhn}, \fref{fig:svhn_cnn_model_a} draws the full architecture: Conv2d (20 filters, kernel size 5x5, stride 1x1), MaxPool2d,  Conv2d (16 filters, kernel size 5x5, stride 1x1), fully-connected 1 (FC1 with 400 neurons), fully-connected 2 (FC2 with 120 neurons), fully-connected 3 (FC3 with 84 neurons); \fref{fig:svhn_cnn_model_b} outlines the \gls{bsssom} training pipeline, where the FC3 is removed, the features are extracted from FC2, and then sent to \gls{bsssom} with 84 input dimensions.

\tref{tab:results_batch} illustrates the best results over 10 runs on each dataset. It showed that, as expected, \gls{bsssom} has increasing gains as the number of labeled samples grows, specifically for beginning percentages. Following through, at a certain point, around 5\% of labeled data, the performance stabilizes. This behavior is observed across all the datasets, showing that the proposed method is a good approach to the problem at hand. Notice that transfer learning is a difficult task, and it is a challenge for a great variety of methods. Such performance obtained by the \gls{bsssom} defines a promising path through the use and application of \gls{som}-based methods.


\begin{table}[ht]
\small
\centering
\renewcommand{\arraystretch}{1.3}
\caption{\textmd{The Accuracy results obtained with \gls{bsssom} on each dataset according to a percentage of labeled data.}}
\label{tab:results_batch}
\centering
\begin{tabular}{c|cccc}
\hline
\bfseries \% & \bfseries MNIST & \bfseries SVHN & \bfseries Fashion-MNIST \\
\hline
1\% & 0.788 & 0.560 & 0.624 \\ 
5\% & 0.9643 & 0.716 & 0.797 \\ 
10\% & 0.974 & 0.713 & 0.798 \\ 
25\% & 0.9793 & 0.777 & 0.834 \\ 
50\% & 0.983 & 0.792 & 0.847 \\ 
75\% & 0.9839 & 0.810 & 0.840 \\ 
All & 0.9836 & 0.826 & 0.846 \\
\hline
\end{tabular}
\end{table}

\section{Conclusion and Future Work}
\label{sec:conclusions}

This paper presented the \gls{bsssom}, an approach that can be applied to both classification and clustering tasks. The proposed model showed a good performance in comparison with other traditional models and also demonstrated its capabilities in the context of having to deal directly with more complex datasets and its representations.

Although the proposed approach is not far superior to other models, it can trace a promising path to follow. It can be considered as the first step towards more SOM-based models that can work effectively in non-traditional scenarios. 


Our main contributions include modifications in the previous model behavior to allow dealing with more complex data structures, while still performing well in traditional tasks for which it was initially intended to do. Finally, for future work, we have left some more detailed studies with transfer learning, optimizations on the  \gls{bsssom} model, and a better way to estimate the unsupervised error when prior information of labels is not given.

\section*{ACKNOWLEDGMENTS}
The authors would like to thank the Brazilian National Council for Technological and Scientific Development (CNPq) and  Coordination for the Improvement of Higher Education Personnel (CAPES) for supporting this research study. Moreover, the authors also gratefully acknowledge the support of NVIDIA Corporation with the GPU Grant of a Titan V.


\bibliographystyle{IEEEtran}
\bibliography{default_content/bibliography}

\end{document}